\documentclass[letterpaper]{article} %
\usepackage{aaai23}  %
\usepackage{times}  %
\usepackage{helvet}  %
\usepackage{courier}  %
\usepackage[hyphens]{url}  %
\usepackage[pdftex]{graphicx} %
\usepackage{natbib}  %
\usepackage{caption} %
\DeclareCaptionStyle{ruled}{labelfont=normalfont,labelsep=colon,strut=off} %
\usepackage[utf8]{inputenc} %
\usepackage[T1]{fontenc}    %
\usepackage{booktabs}       %
\usepackage{amsfonts}       %
\usepackage{nicefrac}       %
\usepackage{microtype}      %
\usepackage{xcolor}         %

\usepackage{tikz}
\usepackage{tikzscale}
\usepackage{subfig}
\usepackage{stackengine}
\usepackage{scalerel}

\usepackage[shortlabels]{enumitem}

\usepackage{bm}
\usepackage{amsthm}
\usepackage{amsmath}
\usepackage{amssymb}

\newtheorem{definition}{Definition}
\newtheorem{theorem}{Theorem}
\newtheorem{corollary}{Corollary}
\theoremstyle{proposition}
\newtheorem{proposition}{Proposition}

\usepackage{algorithm}
\usepackage[noend]{algpseudocode}
\algnewcommand\algorithmicforeach{\textbf{for each}}
\algdef{S}[FOR]{ForEach}[1]{\algorithmicforeach\ #1\ \algorithmicdo}

\newcommand{\indep}{\rotatebox[origin=c]{90}{$\models$}}
\newcommand{\nindep}{\not\!\perp\!\!\!\perp}

\newtheorem{assumption}{A}

\newtheorem{pblm}{Problem}

\newcommand{\ggms}{\text{$GG_{\mathcal{M}_s}$}}

\newcommand{\sagg}{\text{$\sigma$-AGG}}

\newcommand{\mstance}{Sentiment}

\newcommand{\im}[2][\sigma]{\text{$IM_{#1}(#2)$}}
\newcommand{\mr}[2][\sigma]{\text{$\mathcal{P}_{RCD}(IM_{#1}(#2))$}}

\newcommand*\stararrow{%
  \stackengine{0pt}{\hspace{.81ex}$\rightarrow$}{$*$}{O}{l}{F}{F}{L}}

\newcommand*\edgestar{%
  \stackengine{0pt}{\hspace{.81ex}---}{$*$}{O}{l}{F}{F}{L}}

\usepackage{todonotes}

\title{Learning Relational Causal Models with Cycles through Relational Acyclification}
\author{
Ragib Ahsan\textsuperscript{\rm 1},
David Arbour\textsuperscript{\rm 2}, 
Elena Zheleva\textsuperscript{\rm 1}
}
\affiliations{
\textsuperscript{\rm 1}University of Illinois at Chicago,
\textsuperscript{\rm 2}Adobe Research
}

\begin{document}

\maketitle

\begin{abstract} 

In real-world phenomena which involve mutual influence or causal effects between interconnected units, equilibrium states are typically represented with cycles in graphical models. An expressive class of graphical models, \textit{relational causal models}, can represent and reason about complex dynamic systems exhibiting such cycles or feedback loops. 
Existing cyclic causal discovery algorithms for learning causal models from observational data assume that the data instances are independent and identically distributed which makes them unsuitable for relational causal models. At the same time, causal discovery algorithms for relational causal models assume acyclicity. In this work, we examine the necessary and sufficient conditions under which a constraint-based relational causal discovery algorithm is sound and complete for \textit{cyclic relational causal models}. We introduce \textit{relational acyclification}, an operation specifically designed for relational models that enables reasoning about the identifiability of cyclic relational causal models. We show that under the assumptions of relational acyclification and $\sigma$-faithfulness, the relational causal discovery algorithm RCD~\cite{maier-uai13} is sound and complete for cyclic models. We present experimental results to support our claim.

\end{abstract}

\section{Introduction}

Most of the tools and methods developed for causal discovery rely on a graphical representation based on Bayesian networks which assume independent and
identically distributed (i.i.d) instances. Probabilistic relational models~\cite{getoor-book07} have been developed that relax this assumption. The key advantage of these \textit{relational} models is that they can represent systems involving multiple types of entities interacting with each other with some probabilistic dependence. Causal reasoning over such relational systems is key to understanding many real-world phenomena, such as social influence. 

Influence in complex dynamic systems is often mutual and represented by a feedback loop or cycle in the relational model. Identifying mutual influence in relational models is of great interest in the research community. For example, social scientists and marketing experts are interested to study the social dynamics between people and products in social networks~\cite{bakshy-wsdm11,bakshy-science15,ogburn-stats20}. However, there is a lack of available methods for discovering mutual influence or cycles in complex relational systems. 

Sound and complete algorithms have been proposed for learning relational causal models from observational data~\cite{maier-uai13,lee-uai16,lee-aaai16}. However, they assume acyclicity and thus cannot reason about mutual influence or cycles. In a recent work, \citet{ahsan-clear22} develop $\sigma$-abstract ground graph (\sagg{}), a sound and complete representation for cyclic relational causal models. Even though \sagg{} is shown to be sound and complete for cyclic relational causal models, to the best of our knowledge no work on discovering \sagg{} or identifying relational cycles from observational data exists in the literature.

The closest works on cyclic causal discovery are mostly from the domain of Bayesian networks. \citet{richardson-uai96} develop a cyclic causal discovery (CCD) algorithm which is shown to be sound but not complete. In recent work, \citet{mooij-uai20} 
provide necessary conditions for constraint-based causal discovery algorithms developed for acyclic causal models, such as PC~\cite{pearl-book00} and FCI~\cite{spirtes-mitp00}, to be sound and complete for cyclic causal models under sigma-separation criteria. There are several other algorithms for cyclic causal discovery from i.i.d samples~\cite{rothenhausler-nips15,strobl-springer19} but no such algorithm exists for cyclic relational causal models.

In this work, we examine the necessary and sufficient conditions for which constraint-based relational causal discovery can be shown to be sound and complete for cyclic relational causal models under $\sigma$-separation. We introduce \textit{relational acyclification}, an operation that helps to reason over the scope of cyclic relational models which are identifiable with constraint-based causal discovery algorithms. Following this criterion, we show that RCD~\cite{maier-uai13}, a pioneering relational causal discovery algorithm for acyclic relational models, is sound and complete for cyclic relational models under $\sigma$-separation and causal sufficiency assumption. We provide experimental results on synthetic relational models in support of our claims. We also demonstrate the effectiveness of the algorithm on a real-world dataset.

\section{Related Work}

\citet{richardson-uai96} develop cyclic causal discovery (CCD) algorithm for directed cyclic graphs under the causal sufficiency assumption. They provide a characterization of the equivalence class of cyclic causal models. They show that a class of graphs called Partial Ancestral Graphs (PAG) is sufficient to represent the equivalence class of cyclic causal models. Two models are equivalent if they entail the same set of d-separation relationships. There are a few caveats about CCD and its choice of representation. Markov property of $d$-separation holds for directed cyclic graphs only with linear structural equation models (SEM). So, it is not sure how to extend CCD to more general models beyond linear SEMs. Moreover, CCD is not complete in the sense that it does not guarantee to produce the maximally oriented PAG.

\citet{rothenhausler-nips15} develop a general discovery algorithm (BackShift) that allows latent confounders in addition to cycles. The method relies on equilibrium data of the model recorded under a specific kind of intervention called \textit{shift interventions}. \citep{strobl-ijdsa19} develop the CCI algorithm which allows both latent confounders and selection bias apart from cycles. Similar to CCD, both Backshift and CCI are restricted to only linear SEM models. CCI considers a different representation for equivalence class, called maximal almost ancestral graph (MAAG) which is carefully chosen to allow for both latent confounders and selection bias. 

The FCI algorithm is a constraint-based causal discovery algorithm designed specifically for acyclic causal models with latent confounders~\cite{spirtes-mitp00}. \citet{mooij-uai20} show that FCI is sound and complete for cyclic models under $\sigma$-separation criteria which is different than $d$-separation and not restricted to only linear models. They also show that any constraint-based causal discovery algorithm (PC, FCI~\cite{spirtes-mitp00} etc.) which is sound and complete for acyclic causal models, can be shown to be sound and complete for cyclic causal models under some background knowledge (i.e. sufficiency) and assumptions.

\citet{maier-uai13} develop the first sound and complete algorithm that can discover the dependencies of a relational causal model under the assumption of $d$-faithfulness, sufficiency, acyclicity. It is designed based on the PC algorithm with some additional steps introduced specifically to handle relational aspects of the representation. They utilize the \textit{abstract ground graph}, an abstract representation that allows answering relational queries based on $d$-separation criterion~\cite{maier-arxiv13}. RCD introduces Relational Bi-variate Orientation (RBO)- an orientation rule specifically designed for relational models. \citet{lee-aaai16} develop an efficient version of RCD named RCD-Light which requires polynomial time to run. They also develop an alternative algorithm RpCD based on \textit{path semantic} which describes a unique way of defining relational paths~\cite{lee-uai16}.

\citet{ahsan-clear22} develop $\sigma$-abstract ground graph, a sound and complete abstract representation for cyclic relational causal models under $\sigma$-separation. They introduce relational $\sigma$-separation and show that this criterion can consistently answer relational queries on cyclic relational models.

\section{Preliminaries}

We present a comprehensive description of the necessary notation and terminologies used in graphical causal modeling and relational causal models in the Appendix. Here, we present a high-level overview of the important notions used in this paper. For more details, we refer the reader to the literature (\cite{pearl-book00,spirtes-mitp00,richardson-uai96,forre-arxiv17,mooij-uai20,maier-uai13,lee-uai15}).

\subsection{Cyclic Graphical Causal Models}

The most common graphical representation for causal models is \textit{directed acyclic graphs (DAGs)}. 

DAGs provide ways for natural causal interpretation and satisfy the Markov property under $d$-separation. A more general class of graphs are \textit{directed cyclic graphs (DCGs)} which drop the assumption of acyclicity (and allow feedback loops). These graphs are appropriate for (possibly cyclic) structural causal models (SCMs) where the corresponding Markov properties and causal interpretation are more subtle~\cite{bongers-arxiv21}. Cyclic SCMs are useful to represent causal semantics of equilibrium states in dynamical systems~\cite{bongers-arxiv21}.

Directed cyclic graphs offer certain properties that help model cyclic causal models. Given a directed cyclic graph $G = (V,E)$, all nodes on directed cycles passing through node $i \in V$ together form the strongly connected component $SC_G(i) = AN_G(i) \bigcap DE_G(i)$ of $i$ where $AN_G(i)$ and $DE_G(i)$ refers to the ancestors and descendants of node $i \in V$. The set of conditional independence entailed in DCG, $G$ is refereed to as independence model $IM(G)$.

Unlike DAGs, DCGs are not guaranteed to satisfy the Markov property in a general case under $d$-separation. Instead, a more general notion of separation, called $\sigma$-separation satisfies the Markov property of DCGs~\cite{forre-arxiv17}. $\sigma$-separation states that a non-collider blocks a path only if it points to another node in the path which belongs to a different strongly connected component~\cite{mooij-uai20}. $\sigma$-\textit{faithfulness} refers to the property which states that all statistical dependencies found in the distribution generated by a given causal model are entailed by the $\sigma$-separation relationships.

\citet{richardson-uai96} show that a class of graphs called Partial Ancestral Graphs (PAG) is a sufficient representation for the equivalence class of cyclic causal models represented by DCGs. PAGs have also been shown to be a sufficient representation for causal discovery with cycles and unobserved confounders~\cite{mooij-uai20}. Since we are assuming no selection bias for simplicity, we will only discuss directed PAGs (DPAG) in this study.

\citet{forre-arxiv17} introduced an operation called \textit{acyclification} for directed cyclic graphs that generates DAGs with equivalent independence models as the given DCG. It allows a single DPAG to represent the ancestral relationship of a DCG $G$ and all its acyclifications $G^\prime$.

\begin{definition} [Acyclification \cite{forre-arxiv17}]
\label{dfn:acy}
Given a DCG $G = (\mathcal{V}, \mathcal{E})$, an acyclification of $G$ is a DAG $G^{\prime} = (\mathcal{V}, \mathcal{E}^{\prime})$ with

\begin{enumerate}[i]
    \item the same nodes $\mathcal{V}$;
    \item for any pair of nodes ${i, j}$ such that $i \notin SC_G(j)$: $i \rightarrow j \in \mathcal{E}^{\prime}$ iff there exists a node $k$ such that $k \in SC_G(j)$ and $i \rightarrow k \in \mathcal{E}$;
    \item for any pair of distinct nodes ${i, j}$ such that $i \in SC_G(j)$: $i \rightarrow j \in \mathcal{E}^{\prime}$ or $i \leftarrow j \in \mathcal{E}^{\prime}$;
\end{enumerate}
\end{definition}

\begin{proposition}[\cite{mooij-uai20}]
\label{prop:im}
For any DCG $G$ and any acyclification $G^\prime$ of $G$, $IM_\sigma(G) = IM_\sigma(G^\prime) = IM_d(G^\prime)$ where $IM_\sigma(G)$ and $IM_d(G)$ refers to the independence model of the given DCG $G$ under $\sigma$-separation and $d$-separation respectively.
\end{proposition}

\citet{mooij-uai20} provide the necessary conditions under which constraint-based causal discovery algorithms for acyclic causal models, such as PC~\cite{pearl-book00} and FCI~\cite{spirtes-mitp00}, are sound and complete in the presence of cycles under $\sigma$-separation. Their result depends on the following assumptions:

\begin{assumption}
\label{asm:sfaith}
The underlying causal model is $\sigma$-faithful.
\end{assumption}

\begin{assumption}
\label{asm:acy}
There exists one or more valid acyclifications of the given causal model which contains the same set of ancestral relationships as the given model.~\footnote{See Appendix for details}
\end{assumption}

\begin{corollary}[\cite{mooij-uai20}]
\label{cor:pc_gen}
The PC algorithm with Meek's orientation rules is sound, arrowhead-complete, tail-complete, and Markov complete (in the $\sigma$-separation setting without selection bias) for directed cyclic graphs.~\cite{mooij-uai20}
\end{corollary}

\begin{figure}[!ht]
    \centering

    \subfloat{
        \label{sfig:rcm}
        \includegraphics[width=\linewidth]{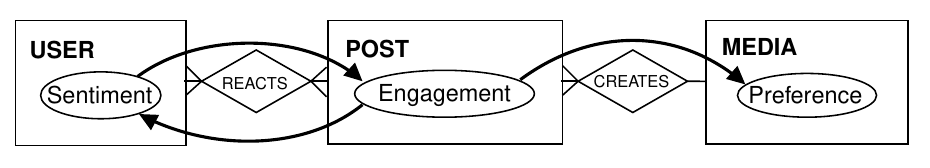}
    }
    
    \caption{
        Example of a cyclic relational model. There are three entity types (USER, POST, MEDIA) and two relationship types (REACTS, CREATES) among them. Attributes are shown in oval shapes. The bold arrows refer to the relational dependencies.
    }
    \label{fig:rcm}
\end{figure}

\begin{figure*}[!ht]
    \centering

    \subfloat[Skeleton]{
        \label{sfig:skel}
        \includegraphics[width=0.25\linewidth]{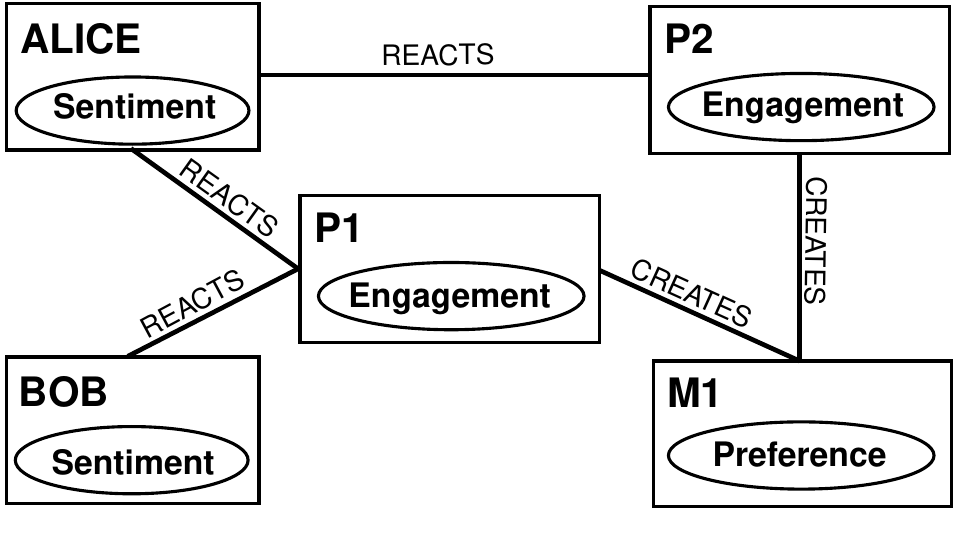}
    }
    \subfloat[Ground Graph]{
        \label{sfig:gg}
        \includegraphics[width=0.25\linewidth]{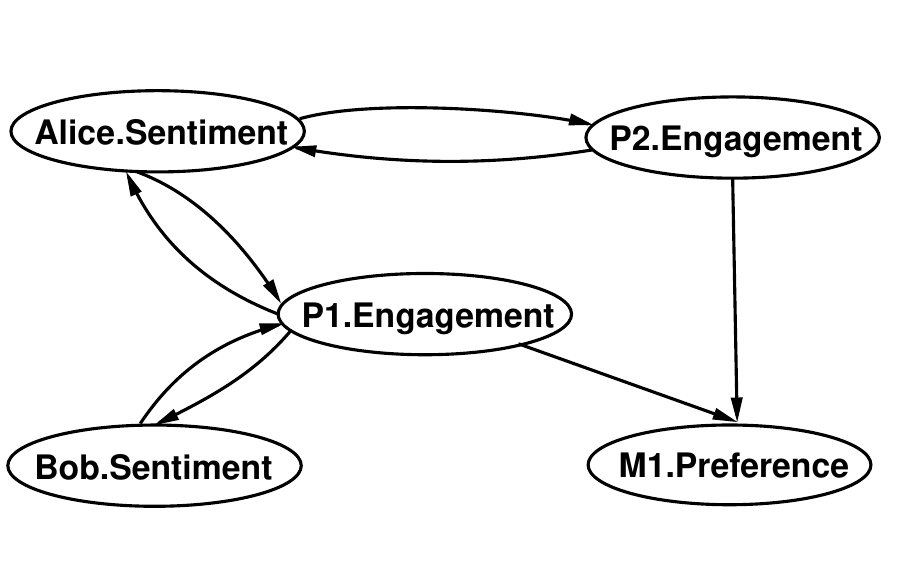}
    }
    \subfloat[Abstract Ground Graph]{
        \label{sfig:agg}
        \includegraphics[width=0.50\linewidth]{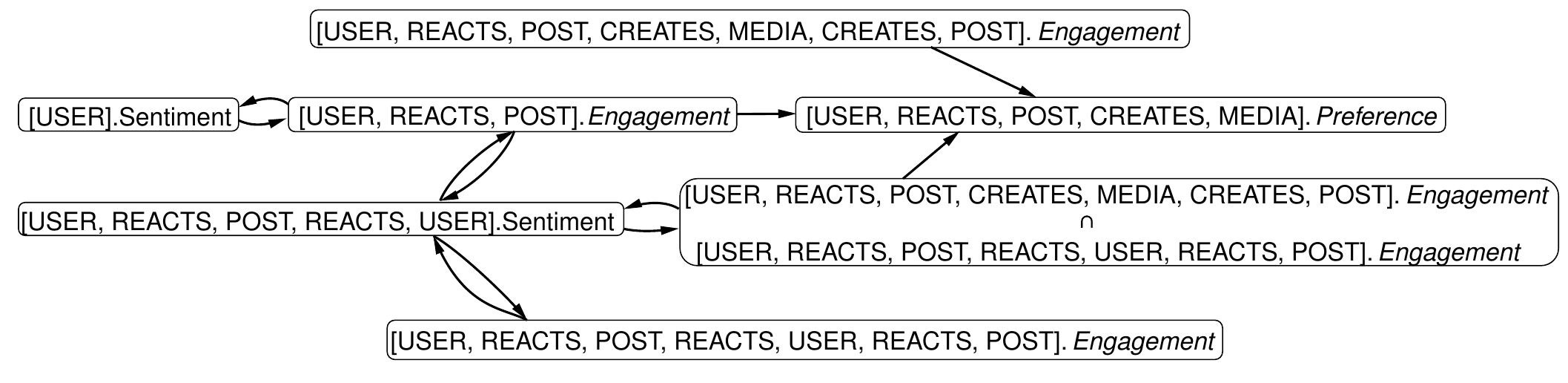}
    }
    
    \caption{
        Fragments of a relational skeleton, ground graph, and $\sigma$-abstract ground graph corresponding to the relational causal model from Figure \ref{fig:rcm}. The arrows represent relational dependencies.
    }
    \label{fig:rcm_all}
\end{figure*}

\subsection{Relational Causal Models (RCMs)}
\label{sec:rcm}

Relational causal models are an expressive class of graphical models that can represent probabilistic dependence among instances of different entity types interacting with each other following a specific schema. We use a simplified Entity-Relationship model to describe relational models following previous work ~\citep{heckerman-isrl07,maier-uai13,lee-uai20}. A relational schema $\mathcal{S} = \langle \bm{\mathcal{E}}, \bm{\mathcal{R}}, \bm{\mathcal{A}}, card \rangle$ represents a relational domain where $\bm{\mathcal{E}}$, $\bm{\mathcal{R}}$ and $\bm{\mathcal{A}}$ refer to the set of entity, relationship and attribute classes respectively. Figure \ref{fig:rcm} shows an example relational model that describes a simplified user-media engagement system. The cardinality constraints are shown with crow’s feet notation— a user can react to multiple posts, multiple users can react to a post, and a post can be created by only a single media entity.

A relational causal model $\mathcal{M} = \langle \mathcal{S},\mathcal{D} \rangle$, is a collection of relational dependencies defined over schema $\mathcal{S}$. \textit{Relational dependencies} consist of two relational variables, cause, and effect. As an example, consider the following relational dependency [Post, Reacts, User].Sentiment $\rightarrow$ [Post].Engagement which states that the engagement of a post is affected by the actions of users who react to that post. Note that, all causal dependencies are defined with respect to a specific \textit{perspective} (entity type). A relational model $\mathcal{M} = (\mathcal{S}, \mathcal{D})$ is said to be \textit{cyclic} if the set of relational dependencies $\mathcal{D}$ construct one or more directed cycles of arbitrary length. There is a direct feedback loop in the relational model of Figure \ref{fig:rcm} making it a cyclic relational causal model.

\sloppy
A realization of a relational model $\mathcal{M}$ with a relational skeleton is referred to as the \textit{ground graph} $GG_\mathcal{M}$. It is a directed graph consisting of attributes of entities in the skeleton as nodes and relational dependencies among them as edges. Figure \ref{sfig:gg} shows the ground graph for the relational model from Figure \ref{fig:rcm}. A $\sigma$-\textit{abstract ground graph} ($\sigma$-AGG) is an abstract representation that captures the dependencies consistent in all possible ground graphs and represents them in a directed graph. $\sigma$-AGGs are defined for a specific perspective and \textit{hop threshold}, $h$. Hop threshold refers to the maximum length of the relational paths. Figure \ref{sfig:agg} presents the $\sigma$-AGG from the perspective of USER with $h = 6$. 

Conditional independence facts are only useful when they hold across all ground graphs that are consistent with the model. \citet{maier-arxiv13} show that relational $d$-separation is sufficient to achieve that for acyclic models. In recent work, \citep{ahsan-clear22} introduced relational $\sigma$-separation criteria specifically for cyclic relational models which directly follows from the definition of relational $d$-separation except it uses $\sigma$-separation criterion instead of $d$-separation.

\subsection{Relational Causal Discovery (RCD)}

The RCD algorithm developed by \citet{maier-uai13} is the first sound and complete algorithm that can discover the dependencies of a relational causal model (RCM) under the assumption of $d$-faithfulness, sufficiency, acyclicity, and a maximum hop threshold $h$. It is designed based on the PC algorithm with some additional steps introduced specifically to handle relational aspects of the representation. \citet{maier-uai13} provides theoretical guarantees for soundness and completeness of RCD.
\section{Relational Causal Discovery with Cycles}

Cyclic relational causal models (CRCM) are relational causal models where dependencies form one or more directed cycles~\citep{ahsan-clear22}. The cycles or feedback loops can represent equilibrium states in dynamic systems. Consider the example from Figure \ref{fig:rcm} where sentiments of users and engagements in a media post may reach an equilibrium. Identifying such cycles or feedback loops from observational samples requires proper representation and a learning algorithm. \citet{ahsan-clear22} introduce an abstract representation, \sagg{} that entails all the conditional independence relations consistent across all ground graphs of the model and shows that it is sound and complete under $\sigma$-separation. Given \sagg{} representation, discovering CRCM transforms into the problem of learning the \sagg{} from observational samples of a relational model. Since \sagg{} is a DCG, we can consider DPAGs to represent the equivalence class of \sagg{} following the previous work of \citet{richardson-uai96}.

\begin{pblm} [Cyclic Relational Causal Discovery]
Given observational samples from a $\sigma$-faithful cyclic relational causal model $\mathcal{M} = \langle \mathcal{S},\mathcal{D} \rangle$ with hop threshold $h$, learn the maximally oriented DPAG that contains the corresponding $\sigma$-AGGs of $\mathcal{M}$.
\end{pblm}

\begin{figure*}[!ht]
    \centering
    \subfloat[Cyclic RCM]{\label{sfig:counter_model}\includegraphics[width=.30\textwidth]{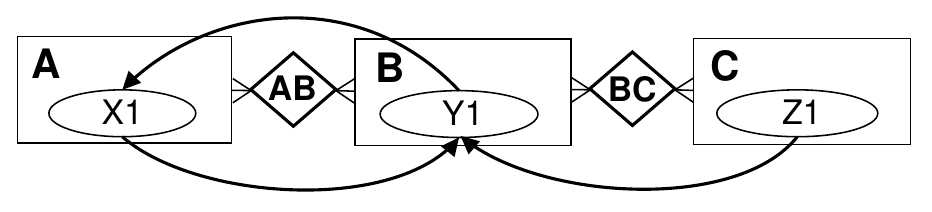}}\hfill
    \subfloat[True $\sigma$-AGG for perspective A]{\label{sfig:true_agg}\includegraphics[width=.30\textwidth]{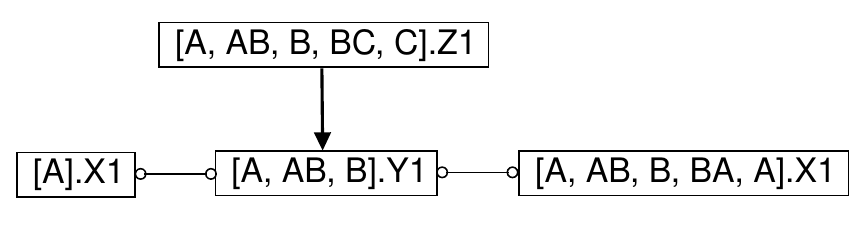}}\hfill
    \subfloat[RCD output for perspective A]{\label{sfig:rcd_agg}\includegraphics[width=.30\textwidth]{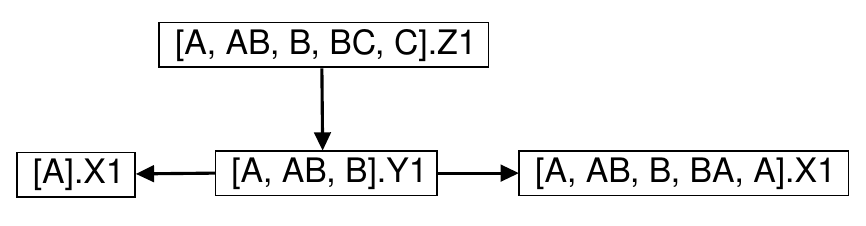}}\\
    
    \caption{Counterexample showing RCD produces incorrect output for cyclic RCM under $\sigma$-separation.
    }
    \label{fig:rcd_counter}
\end{figure*}

\subsection{RCD for cyclic relational causal models}

The RCD algorithm developed by \citet{maier-uai13} is the first sound and complete constraint-based algorithm that can learn relational dependencies of a relational causal model (RCM) under the assumption of $d$-faithfulness, sufficiency, acyclicity, and a maximum hop threshold $h$. It is designed based on the PC algorithm with additional steps introduced specifically to handle relational aspects of the representation.

Following the recent development by \citet{mooij-uai20} (Corollary \ref{cor:pc_gen}), and considering that RCD is developed based on the PC algorithm, a natural question arises: \textit{Is RCD sound and complete for cyclic relational causal models?} To the best of our knowledge, no prior work addresses this question. More generally, the effectiveness and theoretical guarantees of existing relational causal structure learning algorithms for cyclic RCMs under $\sigma$-separation are not studied in the current literature.

\subsubsection{A counterexample}
We present a counterexample that shows that RCD is not sound and complete for discovering cyclic relational causal models in general. Figure \ref{sfig:counter_model} shows a CRCM with three entity types A,B,C, and two relationship types AB, BC and
maximum hop threshold $h = 2$. The attribute types X1, Y1, and Z1 refer to the attributes of entity types A, B, and C respectively. There are three relational dependencies: 1) [A, AB, B].Y1 $\rightarrow$ [A].X1, 2) [B, AB, A].X1 $\rightarrow$ [B].Y1, and 3) [B, BC, C].Z1 $\rightarrow$ [B].Y1. The first two dependencies form a feedback loop. Figure \ref{sfig:true_agg} shows the true \sagg{} built from perspective A with maximum hop threshold $h = 4$~\footnote{A sound and complete AGG contains nodes with higher hop threshold than the model($h$)~\cite{maier-arxiv13}. However, the hop threshold of relational dependencies cannot exceed $h$. The same argument holds for $\sigma$-AGGs as well.}. Figure \ref{sfig:rcd_agg} shows the output of RCD with a $\sigma$-separation oracle. We see that RCD orients arrows [A, AB, B].Y1 $\rightarrow$ [A].X1 and [A, AB, B].Y1 $\rightarrow$ [A, AB, B, AB, A].X1 which refers to the relational dependency [A, AB, B].Y1 $\rightarrow$ [A].X1. However, the true model contains a feedback loop between [A, AB, B].Y1 and [A].X1. This example shows that RCD, even with $\sigma$-separation oracle produces incorrect edge orientations.

\section{Relational Acyclification for Cyclic Relational Causal Models}

In this section, we present relational acyclification which enables the discovery of relational causal models with cycles. We also discuss how to read off features of the true model from the output of the discovery algorithm.

\subsection{Relational Acyclification}

The counterexample in the previous subsection shows that the RCD algorithm is not sound and complete for general cyclic RCMs under $\sigma$-separation. For the given counterexample, RCD orients edges that contradict the given relational model. In order to understand what causes this error and to find a solution, we focus on the acyclification operation introduced by \citet{forre-arxiv17} which is a key tool for the generalization results by \citet{mooij-uai20}. 

\begin{figure}[ht!]
    \centering
    \includegraphics[width=.30\textwidth]{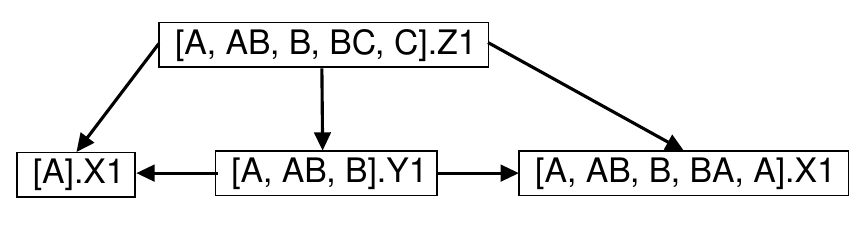}
    
    \caption{Invalid acyclification of \sagg{} from Figure \ref{sfig:true_agg}}
    \label{fig:acy}
\end{figure}

Figure \ref{fig:acy} shows an acyclification of the \sagg{} presented in Figure \ref{sfig:true_agg} following definition \ref{dfn:acy}. Here we see the edges [A, AB, B, BC, C].Z1 $\rightarrow$ [A].X1 and [A, AB, B, BC, C].Z1 $\rightarrow$ [A, AB, B, AB, A].X1 which does not follow the relational model since the hop threshold of such dependencies ($h$ = 4) exceed the hop threshold of the given model ($h$ = 2). The definition of acyclification, as given by \cite{forre-arxiv17} essentially considers all the nodes or entities to be of the same entity type. As a result, applying it directly to relational models creates erroneous results. We propose a new definition of acyclification for relational models which specifically mentions that the maximum hop threshold of an acyclification can be different than the hop threshold of the original model.

\begin{definition} [Relational Acyclification]
\label{dfn:rel_acy}
Given a relational schema $\mathcal{S} = (\mathcal{E}, \mathcal{R}, \mathcal{A}, card)$, $\sigma$-AGG $G = (V, E)$, and a hop threshold $h$, a relational acyclification of $G$ is a $\sigma$-AGG $G^{\prime} = (V, E^{\prime})$ with hop threshold $h^\prime \ge h$  containing

\begin{enumerate}[i]
    \item the same nodes $V$;
    \item for any pair of nodes ${P.X, Q.Y}$ such that $P.X \notin SC_G(Q.Y)$: $P.X \rightarrow Q.Y \in E^{\prime}$ iff there exists a node $R.Z$ such that $R.Z \in SC_G(Q.Y)$ and $P.X \rightarrow R.Z \in E$ and $P.X \rightarrow Q.Y$ is a valid relational dependency with maximum hop threshold $h^\prime$;
    \item for any pair of distinct nodes ${P.X, Q.Y}$ such that $P.X \in SC_G(Q.Y)$: $P.X \rightarrow Q.Y \in E^{\prime}$ or $P.X \leftarrow Q.Y \in E^{\prime}$;
\end{enumerate}
\end{definition}

The definition of relational acyclification follows from Definition \ref{dfn:acy} where the main distinction is that it allows a new bound on the maximum hop threshold which is different than the bound of the original model. The implication of this is that the potential dependencies RCD considers in building the skeleton, may not be sufficient for soundness and completeness.

\subsection{Maximum hop threshold for relational acyclification}

Definition \ref{dfn:rel_acy} suggests that the maximum hop threshold used in a relational acyclification of a $\sigma$-AGG may be higher than the hop threshold of the given model. It is important to characterize the maximum bound of relational acyclifications for allowing practical implementation of the RCD algorithm for cyclic models. The following proposition provides the maximum bound on the hop threshold of relational acyclifications.

\begin{proposition}
\label{prop:rel_hop}
Given a relational model $\mathcal{M} = (\mathcal{S}, \mathcal{D})$ with hop threshold $h$ and corresponding \sagg{} $G = (V, E)$ with a given perspective, the hop threshold $h^\prime$ of any relational acyclification $G^\prime$ of $G$ can be at most $\lfloor \frac{2 + l^c}{2} \rfloor h$ where $l^c$ refers to the length of the longest cycle of dependencies in the relational model $\mathcal{M}$.
\end{proposition}

The need for higher hop thresholds arises for the additional edges drawn for any incoming edges to a strongly connected component (Definition \ref{dfn:acy}). Any such incoming edge has a maximum hop threshold $h$ of the given model. In order to reach the farthest node in the cycle where each dependency can be of at most $h$ hop threshold, we need at most $\lfloor \frac{l^c}{2} \rfloor h$ hop threshold where $l^c$ refers to the length of the cycle. So, in total it can be at most $h + \lfloor \frac{l^c}{2} \rfloor h = \lfloor \frac{2 + l^c}{2} \rfloor h$. Note that in order to calculate an upper bound on the hop threshold of relational acyclification we need to assume the maximum length of any cycle, $l^c$ in the given relational model.

\subsection{Soundness and completeness of RCD for cyclic relational causal models}

We consider RCD as a mapping $\mathcal{P}_{RCD}$ from independence models (on variables $V$) to DPAGs (with vertex set $V$), which maps the independence model of a \sagg{} $G$ to the DPAG $\mathcal{P}_{RCD}(IM_\sigma(G))$. We assume the following:

\begin{assumption}
\label{asm:rel_acy}
There exists one or more valid relational acyclifications with hop threshold not exceeding the hop threshold of the given relational causal model ($h^\prime = h$).
\end{assumption}

\begin{assumption}
\label{asm:sagg}
The degree of any entity in the relational skeleton is greater than one.
\end{assumption}

Assumption \ref{asm:rel_acy} follows from Assumption \ref{asm:acy} and also limits the set of relational causal models for which RCD can be shown to be sound and complete. Assumption \ref{asm:sagg} satisfies the soundness and completeness of \sagg{}~\cite{ahsan-clear22}.

\begin{theorem}
Considering Assumption \ref{asm:sfaith}, \ref{asm:rel_acy}, \ref{asm:sagg} and causal sufficiency holds, RCD is 

\begin{enumerate}[(i)]
    \item sound: for all \sagg{}s $G$, \mr{G} contains $G$;
    \item arrowhead complete: for all \sagg{}s $G$: if $i \notin AN_{\tilde{G}}(j)$ for any DCG $\tilde{G}$ that is $\sigma$-Markov equivalent to $G$, then there is an arrowhead $j\; \stararrow \;i$ in \mr{G} 
    \item tail complete: for all \sagg{}s $G$, if $i \in AN_{\tilde{G}}(j)$ 
    in any DCG $\tilde{G}$ that is $\sigma$-Markov equivalent to $G$, 
    then there is a tail $i \rightarrow j$ in \mr{G};
    \item Markov complete: for all \sagg{}s $G_1$ and $G_2$, $G_1$ is $\sigma$-Markov equivalent to $G_2$ iff $\mr{G_1} = \mr{G_2}$
\end{enumerate}
in the $\sigma$-separation setting given sufficient hop threshold.
\end{theorem}

\begin{proof}
    The main idea of the proof is very similar to the proof of Theorem 1 from \citet{mooij-uai20} where they prove the soundness and completeness of FCI for cyclic models under $\sigma$-separation. 
    
    To prove soundness, let $G$ be a \sagg{} and $\mathcal{P} = \mr{G}$. The acyclic soundness of RCD means that for all AGGs $G^\prime$, $\mr{G^\prime}$ contains $G^\prime$. Hence, by Definition \ref{dfn:rel_acy} and Assumption \ref{asm:rel_acy}, $\mathcal{P}$ contains $G^\prime$ for all acyclifications $G^\prime$. But then $\mathcal{P}$ must contain $G$ which can be easily shown using Proposition 3 of \citet{mooij-uai20}.
    
    To prove arrowhead completeness, let $G$ be a \sagg{} and suppose that $i \notin AN_{\tilde{G}}(j)$ in any DCG $\tilde{G}$ that is $\sigma$-Markov equivalent to $G$. Since $G^\prime$ is $\sigma$-Markov equivalent to $G$, this implies in particular that for all AGGs $\tilde{G}$ that are $d$-Markov equivalent to $G^\prime$, $i \notin AN_{\tilde{G}}(j)$. Because of the acyclic arrowhead completeness of RCD, there must be an arrowhead $j\; \stararrow \;i$ in $ \mr{G^\prime} = \mr{G}$. Tail completeness is proved similarly.
    
    To prove Markov completeness: Definition \ref{dfn:rel_acy} and Proposition \ref{prop:im} imply both $\im{G_1} = \im[d]{G_1^\prime}$ and $\im{G_2} = \im[d]{G_2^\prime}$. From the acyclic Markov completeness of RCD\footnote{Since relational d-separation is sound and complete for AGG~\cite{maier-arxiv13}}, it then follows that $G_1^\prime$ must be $d$-Markov equivalent to $G_2^\prime$, and hence $G_1$ must be $\sigma$-Markov equivalent to $G_2$.
\end{proof}

The statement of this theorem can be seen as a special case of the generalization claim (Theorem 2) by \citet{mooij-uai20}. There is an important point to discuss about Assumption \ref{asm:rel_acy}. Even though Assumption \ref{asm:rel_acy} limits the scope of possible relational causal models, it is possible to modify RCD in a way so that it can work for models with relational acyclification having hop threshold higher than the hop threshold of the given model ($h^\prime > h$). The intuition here is that the skeleton building process should consider this new hop threshold $h^\prime$ (which is upper bounded by $\lfloor \frac{2 + l^c}{2} \rfloor h$) rather than the true hop threshold $h$. However, it requires further proof of soundness and completeness with this modified skeleton. We leave this for future work.

\subsection{Identification of relational (non-)cycles}

\citet{mooij-uai20} show that the patterns in strongly connected components in DCGs can be used as a sufficient condition for identifying the absence of certain cyclic causal relations in a complete DPAG. Given Definition \ref{dfn:rel_acy}, the same condition holds for relational models and \sagg{}s as well. We present the necessary and sufficient conditions for identifying non-cycles in the output of RCD following Proposition 10 by \cite{mooij-uai20}:

\begin{proposition}
\label{prop:id}
Let $G$ be a \sagg{} and denote by $\mathcal{P} = \mr{G}$ the corresponding complete DPAG output by RCD. Let $i \neq j$ be two nodes in $\mathcal{P}$. If there is an edge $i\; \circ$---$\circ \;j$ in $\mathcal{P}$, 
and all nodes $k$ for which $k\; \stararrow \;i$ is in $\mathcal{P}$ also have an edge of the same 
type $k\; \stararrow \;j$ (i.e., the two edge marks at $k$ are the same) in $\mathcal{P}$, then there exists a DCG $\tilde{G}$ with $j \in SC_{\tilde{G}}(i)$ that is $\sigma$-Markov equivalent to $G$, but also a DCG $H$ with $j \notin SC_H(i)$ that is $\sigma$-Markov equivalent equivalent to $G$.
\end{proposition}

\begin{figure}
    \centering
    
    \subfloat[Cyclic RCM]{\label{sfig:id_crcm}\includegraphics[width=.17\textwidth]{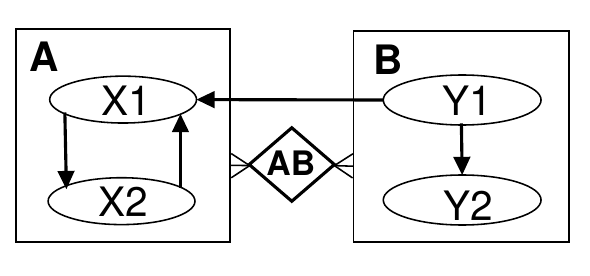}}\hfill
    \subfloat[RCD output DPAG]{\label{sfig:id_pag}\includegraphics[width=.29\textwidth]{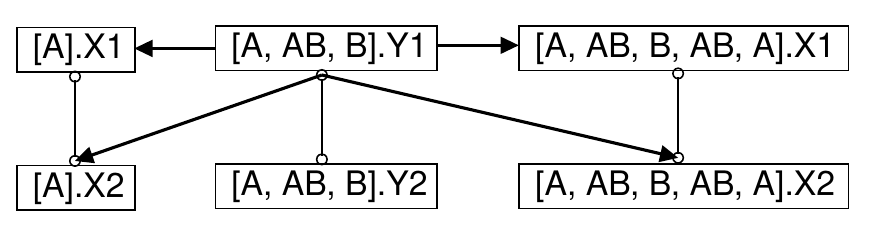}}\hfill
    
    \caption{An example cyclic relational model and its corresponding DPAG output by RCD under $\sigma$-separation.}
    \label{fig:id}
\end{figure}

In other words, under the conditions of this proposition, it is not identifiable from $\mathcal{P}$ alone whether $j$ and $i$ are part of a causal cycle, but they are candidates of being part of a cycle. Figure \ref{fig:id} shows an example of this identifiability criteria. Figure \ref{sfig:id_pag} shows the output DPAG of an example cyclic RCM from Figure \ref{sfig:id_crcm}. The edges between nodes [A].X1, [A].X2 and [A, AB, B, AB, A].X1, [A, AB, B, AB, A].X2 satisfies the conditions given in Proposition \ref{prop:id}. It means they could be part of a cycle but it is not possible to confirm that based on the output alone. 

\section{Experiments}

In this section, we examine the effectiveness of RCD for cyclic RCMs using both synthetically generated cyclic RCMs satisfying relational acyclification criteria and a demonstration with a real-world dataset. Since there is no other algorithm designed to discover cyclic RCMs, we compare against the vanilla RCD with $d$-separation oracle.

\subsection{Experimental Setup}

We follow the procedure introduced by \citet{maier-uai13} for synthetic data generation except we allow feedback loops. We generate 100 random cyclic relational causal models over randomly generated schema for each of the following combinations: entities (1–3); relationships (one less than the number of entities) with cardinalities selected uniformly at random; attributes per item drawn from Pois($\lambda$ = 1) + 1; and the number of relational dependencies (4, 6, 8, 10, 12) limited by a hop threshold of 2 and at most 3 parents per variable. Moreover, for a consistent evaluation, we select models with maximum density of $\sigma$-AGGs within a specific range (0.05 to 0.1 for multiple entity types and 0.3 to 0.5 for single entity type). We enforce a feedback loop among the dependencies. Note that a single feedback loop can introduce arbitrary length cycles based on the structure of the model. This procedure yields a total of 15,000 synthetic models. We refer to the version of RCD with $d$-separation and $\sigma$-separation oracles as $d$-RCD and $\sigma$-RCD respectively.\footnote{Code available at https://github.com/edgeslab/sRCD}

\subsection{Evaluation}

The goal of the evaluation is to compare the learned causal models with the true causal models. However, the output object for cyclic RCMs is PAGs instead of CPDAGs. Moreover, it is expected that the skeleton of the output PAG might be different from the true causal model. For this reason, we evaluate the algorithms based on the ancestral relationships. We identify the ancestral relationships entailed by the output and the \sagg{} of the true model and report F1-score for comparison. For a sound and complete algorithm, we expect to see perfect F1-scores. Moreover, we consider the identification criterion given in Proposition \ref{prop:id} and evaluate the algorithms based on their ability to correctly identify edges as possible cycle candidates. We report F1-score for this evaluation as well.

\subsection{Results}

\begin{figure}[!ht]
    \centering

    \includegraphics[width=.40\textwidth]{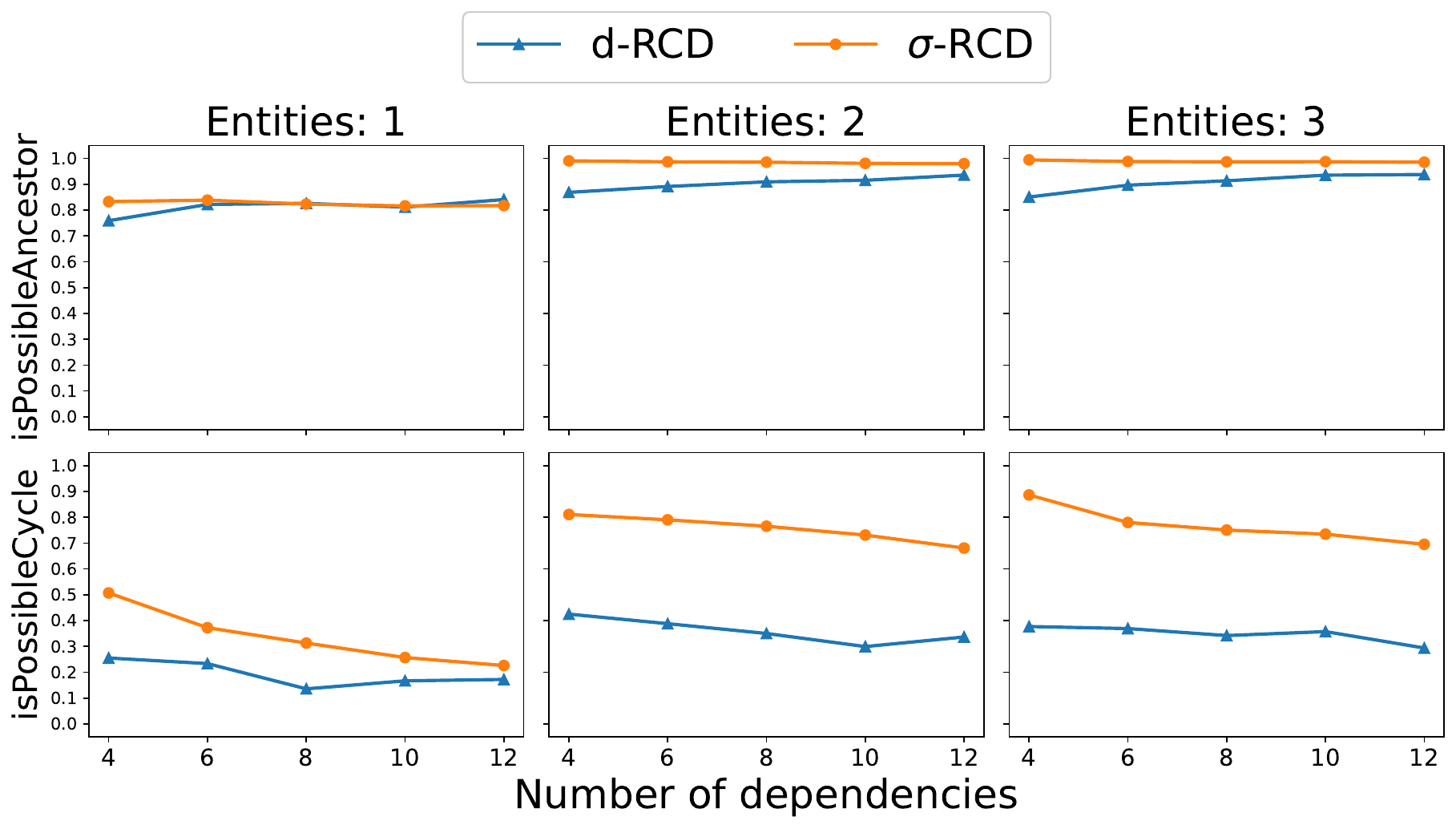}
    
    \caption{
        Comparison of F1-score for $d$-RCD and $\sigma$-RCD based on \textit{isPossibleAncestor} (top row) and \textit{isPossibleCycle} (bottom row) queries. The number of entity types increases from left to right.
     }
    \label{fig:recalls}
\end{figure}

Figure \ref{fig:recalls} shows the comparison of $d$-RCD and $\sigma$-RCD based on \textit{isPossibleAncestor} (top row) and \textit{isPossibleCycle} (bottom row) queries on synthetically generated relational models. The columns represent the increased number of entity types (left to right). The x-axis shows the number of dependencies and y-axis shows F1-scores. In the leftmost column, we see the results for single entity models. The top left and bottom left figures are equivalent to running the PC algorithm with $d$- and $\sigma$-separation oracles respectively. The rest of the figures represent proper relational models. 
In the top row, we see perfect F1-score for $\sigma$-RCD for relational models. However, for single entity models, F1-score is slightly lower than $d$-RCD. However, we recorded 100\% recall for $\sigma$-RCD. We see a general upward trend from left to right which is intuitive since higher number of dependencies make the models increasingly denser. On the other hand, for isPossibleCycle query the trend is generally downward since denser models make the identification harder.

\begin{figure}[!ht]
    \centering

    \includegraphics[width=.40\textwidth]{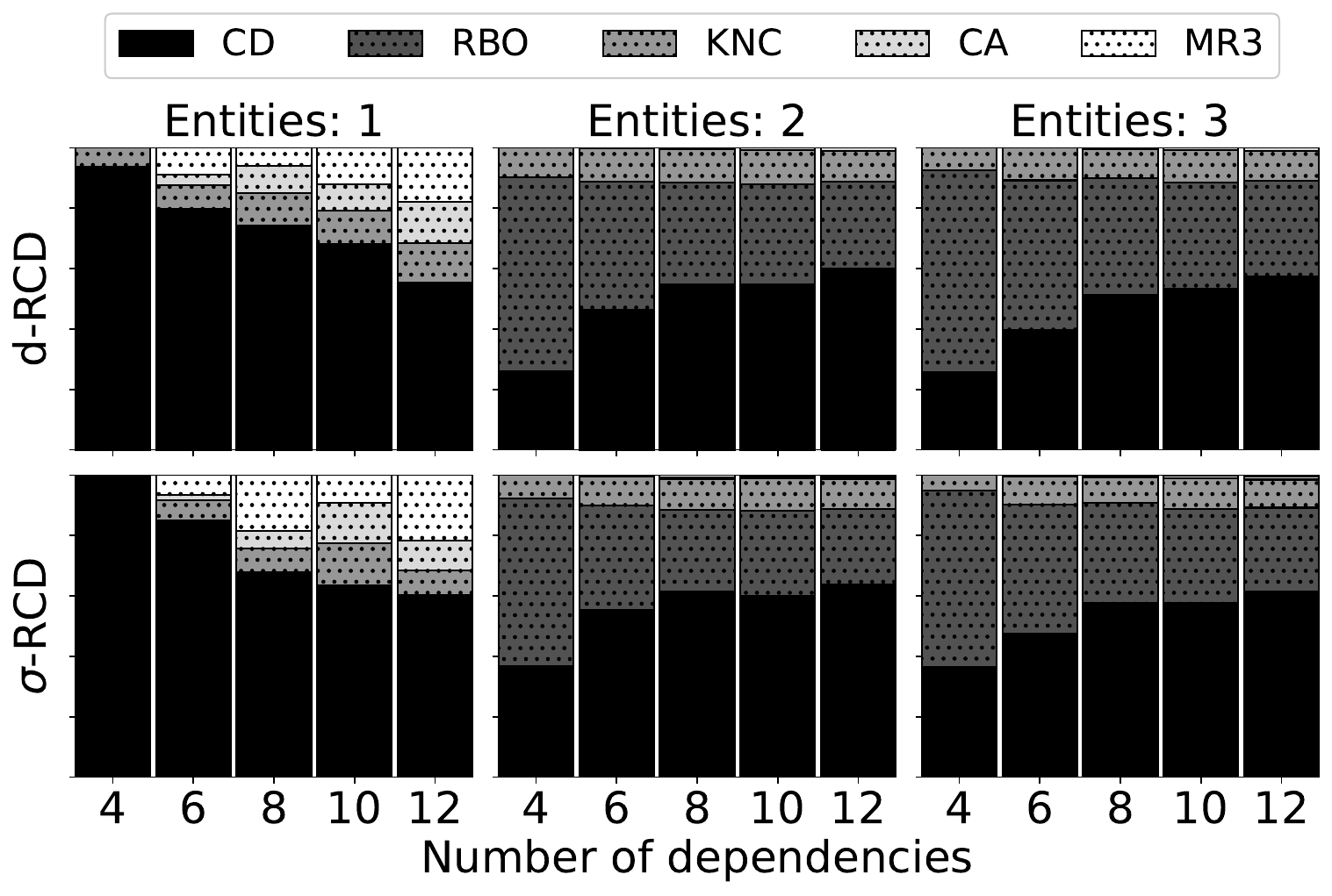}
    
    \caption{
        Frequency of edge orientation rules for $d$-RCD (top) and $\sigma$-RCD for different numbers of entity types and dependencies.
     }
    \label{fig:rules}
\end{figure}

Figure \ref{fig:rules} shows the percentage of orientation rules used for $d$-RCD (top row) and $\sigma$-RCD (bottom row). The leftmost column refers to the single entity case where no RBO is in effect. We can see some subtle differences in the distribution of rules for $d$-RCD and $\sigma$-RCD. For the small number of dependencies (i.e. 4) only CD (collider detection) rule activates with $\sigma$-RCD where $d$-RCD utilizes both CD and KNC (known non-collider). The increased number of dependencies shows the difference in the overall distribution. For the middle and right column, a significant difference is seen in the percentage of times rule MR3 (Meek rule 3) is executed for $\sigma$-RCD compared to $d$-RCD. These differences indicate that the algorithms learn fundamentally different structures.

\subsection{Demonstration on Real-world Data}

\begin{figure}[ht]
    \centering

    \includegraphics[width=.40\textwidth]{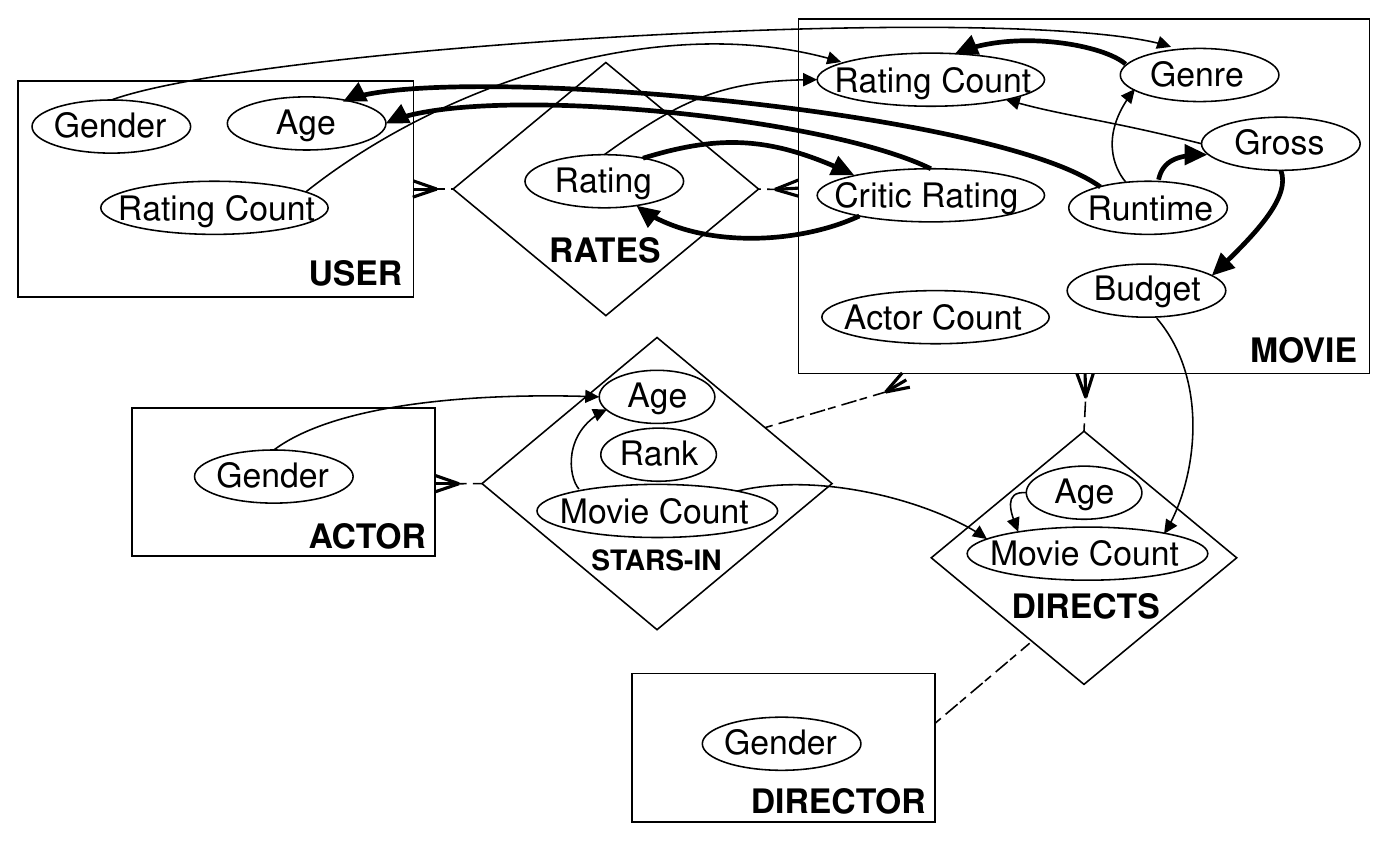}
    
    \caption{
        A possible cyclic relational model of MovieLens+ based on the output of RCD~\cite{maier-uai13}.
     }
    \label{fig:demo}
\end{figure}

\citet{maier-uai13} show the output of RCD on a sample of MovieLens dataset (www.grouplens.org) based on an approximate conditional independence test using the significance of coefficients in linear regressions \footnote{The original output is given in the Appendix}. Their output contains undirected edges which are potential candidates for cycle edges. Figure \ref{fig:demo} shows a possible cyclic relational model which corresponds to the original output. Following Proposition \ref{prop:id}, we can infer that the edge between \textit{[Movie].Rating Count} and \textit{[Movie].Genre} cannot be part of any cycles or feedback loops. Some undirected edges can be oriented based on domain knowledge (i.e. Budget can cause gross income but not the other way around). There exist many possible orientations of dependencies that agrees with the RCD output. We show one plausible case with a feedback loop between \textit{user rating} and \textit{critic ratings} of a movie. It is possible that rating information is public and users and critics influence each other with their ratings. However, it is also possible that there exists one or more unobserved confounders which influence both \textit{user ratings} and \textit{critic ratings}.
\section{Conclusion}

Despite several methods developed for cyclic causal discovery from i.i.d samples, no such algorithm exists for cyclic relational causal models even though cycles are ubiquitous in real-world relational systems. In this work, we investigate the necessary conditions for discovering cyclic relational causal models from observational samples. We introduce relational acyclification operation which facilitates the theoretical guarantees for identifiability of such models. We prove that an existing state-of-the-art relational discovery algorithm, RCD is sound and complete for cyclic relational models for which valid relational acyclification exists. To the best of our knowledge, this discovery is the first of its kind. We hope that this work will play an important role in the study of mutual influence and interference in complex relational systems.

\section{Acknowledgments}

This material is based on research sponsored in part by NSF under grant No.\@ 2047899, and DARPA under contract number HR001121C0168.

\bibliography{references}

\clearpage
\clearpage
\appendix
\label{appendix}

\section{Graphs}

\subsection{Directed Cyclic Graphs}
A Directed Cyclic Graph (DCG) is a graph $\mathcal{G} = \langle \mathcal{V}, \mathcal{E} \rangle$ with nodes $\mathcal{V}$ and edges $\mathcal{E} \subseteq \{(u, v) : u, v \in V, u \neq v\}$ where $(u,v)$ is an ordered pair of nodes. We will denote a directed edge $(u, v) \in \mathcal{E}$ as $u \rightarrow v$ or $v \leftarrow u$, and call $u$ a parent of $v$. In this work, we restrict ourselves to DCG as the causal graphical model. A walk between two nodes $u, v \! \in \! \mathcal{V}$ is a tuple $\langle v_0, e_1, v_1, e_2, v_2, . . . , e_n, v_n \rangle$ of alternating nodes and edges in $\mathcal{G} (n \geq 0)$, such that $v_0, . . . , v_n \in \mathcal{V}$, and $e_1, . . . , e_n \in \mathcal{E}$, starting with node $v_0 = u$ and ending with node $v_n = v$ where the edge $e_k$ connects the two nodes $v_{k-1}$ and $v_k \in \mathcal{G}$ for all $k = 1, . . . , n$. If the walk contains each node at most once, it is called a \textit{path}. A \textit{directed walk (path)} from $v_i \in \mathcal{V}$ to $v_j \in \mathcal{V}$ is a walk (path) between $v_i$ and $v_j$ 
such that every edge $e_k$ on the walk (path) is of the form $v_{k - 1} \rightarrow v_k$, i.e., every edge is directed and points away from $v_i$.

\sloppy
We get the \textit{ancestors} of node $v_j$ by repeatedly following the path(s) through the parents: $AN_\mathcal{G}(v_j) := \{v_i \in V : v_i = v_0 \rightarrow v_1 \rightarrow . . . \rightarrow v_n = v_j \in \mathcal{G}\}$. Similarly, we define the \textit{descendants} of $v_i: DE_\mathcal{G}(v_i) := {v_j \in \mathcal{V} : v_i = v_0 \rightarrow v_1 \rightarrow . . . \rightarrow v_n = v_j \in \mathcal{G}}$. Each node is an ancestor and descendant of itself. 
A directed cycle is a directed path from $v_i$ to $v_j$ such that in addition, $v_j \rightarrow v_i \in \mathcal{E}$. All nodes on directed cycles passing through $v_i \in \mathcal{V}$ together form the strongly connected component $SC_\mathcal{G}(v_i) := AN_\mathcal{G}(v_i) \cap DE_\mathcal{G}(v_i)$ of $v_i$.

\begin{definition} [Strongly connected component (SC)]
Given a directed cyclic graph $G = (V,E)$, all nodes on directed cycles passing through node $i \in V$ together form the strongly connected component $SC_G(i) = AN_G(i) \bigcap DE_G(i)$ of $i$.
\end{definition}

\begin{definition} [Acyclification \cite{forre-arxiv17}]
Given a DCG $G = (\mathcal{V}, \mathcal{E})$, an acyclification of $G$ is a DAG $G^{\prime} = (\mathcal{V}, \mathcal{E}^{\prime})$ with

\begin{enumerate}[i]
    \item the same nodes $\mathcal{V}$;
    \item for any pair of nodes ${i, j}$ such that $i \notin SC_G(j)$: $i \rightarrow j \in \mathcal{E}^{\prime}$ iff there exists a node $k$ such that $k \in SC_G(j)$ and $i \rightarrow k \in \mathcal{E}$;
    \item for any pair of distinct nodes ${i, j}$ such that $i \in SC_G(j)$: $i \rightarrow j \in \mathcal{E}^{\prime}$ or $i \leftarrow j \in \mathcal{E}^{\prime}$;
\end{enumerate}
\end{definition}

\citet{mooij-uai20} presents some important properties of acyclifications:

\begin{proposition}
\label{prop:acy}
Let $G$ be a DCG and $i$, $j$ two nodes in $G$. 

\begin{enumerate}[i]
    \item If $i \in AN_G(j)$ then there exists an acyclification $G^\prime$ of $G$ with $i \in AN_{G^\prime}(j)$;
    \item If $i \notin AN_G(j)$ then $i \notin AN_{G\prime}(j)$ for all acyclifications $G^\prime$ of $G$;
    \item There is an inducing path between $i$ and $j$ in $G$ if and only if there is an inducing path between $i$ and $j$ in $G^\prime$ for any acyclification $G^\prime$ of $G$~\footnote{Inducing paths are special kinds of paths in cyclic graphs. We refer readers to Section 3.1 of \citet{mooij-uai20} for further details.}.
\end{enumerate}
\end{proposition}

\begin{definition} [Independence Model \cite{mooij-uai20}]
An independence model of an DCG $\mathcal{H}$ is given by
\[
    IM_d(\mathcal{H}) := \{ \langle A, B, C \rangle : A, B, C \subset \mathcal{V}, A \overset{d}{\underset{\mathcal{H}}{\indep}} B | C \}
\]
Similarly, a $\sigma$-independence model is defined as
\[
    IM_\sigma(\mathcal{H}) := \{ \langle A, B, C \rangle : A, B, C \subset \mathcal{V}, A \overset{\sigma}{\underset{\mathcal{H}}{\indep}} B | C \}
\]
\end{definition}

The following assumption is an elaboration of Assumption \ref{asm:acy}that states that given background knowledge is \textit{compatible with acyclification} where $\Psi(G) = 1$ refers to the fact that the given background knowledge (i.e. sufficiency) holds \cite{mooij-uai20}.

\begin{assumption}
\label{asm:compat}
For all DCGs $G$ with $\Psi(G) = 1$, the following three conditions hold:
\begin{enumerate}[i]
    \item There exists an acyclification $G^\prime$ of $G$ with $\Psi(G^\prime) = 1$;
    \item For all nodes $i, j \in G$: if $i \in AN_G(j)$ then there exists an acyclification $G^\prime$ of $G$ with $\Psi(G^\prime) = 1$ such that $i \in AN_{G^\prime}(j)$;
    \item For all nodes $i, j \in G$: if $i \notin AN_G(j)$ then there exists an acyclification $G^\prime$ of $G$ with $\Psi(G^\prime) = 1$ such that $i \notin AN_{G^\prime}(j)$;
\end{enumerate}
\end{assumption}

\subsection{PAGs as equivalence class}

\begin{figure}[ht]
    \centering
    \subfloat[DCG $\mathcal{G}$]{\label{sfig:dcg}\includegraphics[width=.18\textwidth]{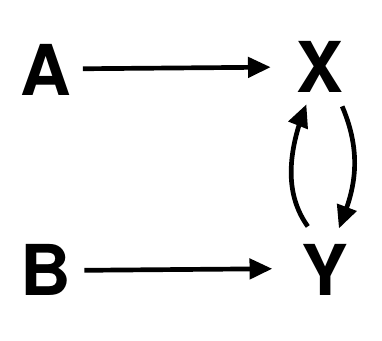}}\hfill
    \subfloat[PAG]{\label{sfig:pag}\includegraphics[width=.18\textwidth]{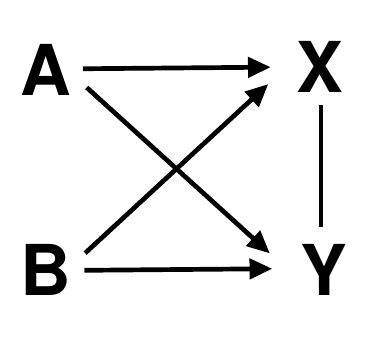}}
    
    \caption{A simple DCG with a feedback loop (left) and its equivalence class (right) }
    \label{fig:ccd}
\end{figure}

The Cyclic Causal Discovery (CCD) algorithm, introduced by \citet{richardson-uai96} is one of the earliest causal discovery algorithms that do not assume acyclicity. They show that a class of graphs called Partial Ancestral Graphs (PAG) is a sufficient representation for the equivalence class of cyclic causal models represented by DCGs. PAGs have also been shown to be a sufficient representation for causal discovery with cycles and unobserved confounders~\cite{mooij-uai20}. Since we are assuming no selection bias for simplicity, we will only discuss directed PAGs (DPAG) in this study. Note that, DPAGs can only represent some features (ancestral relationships) of the true causal model and cannot identify the model itself~\cite{richardson-uai96}. Figure \ref{fig:ccd} shows a simplistic causal model with a feedback loop and its corresponding equivalence class depicted by a DPAG. From Figure \ref{sfig:pag} we can read that both A and B are ancestors of X and Y but X and Y can not be ancestors of A or B. Some of the key features of a DPAG $\Psi$:

\begin{enumerate}
    \item There is an edge between A and B in $\Psi$ iff A and B are connected in DCG $\mathcal{G}$.
    \item If there is an edge in $\Psi$, $B \edgestar A$, out of A (not necessarily into B), then in every graph in Equiv($\mathcal{G}$), A is an Ancestor of B.
    \item If there is an edge in $\Psi$, $A \stararrow B$, into B, then in every graph in Equiv($\mathcal{G}$), B is not an Ancestor of A.
\end{enumerate}

\subsection{$\sigma$-separation}

The idea of $\sigma$-separation follows from $d$-separation, a fundamental notion in DAGs which was first introduced by~\citet{pearl-book88}. $d$-separation exhibits the global Markov property in DAGs which states that if two variables $X$ and $Y$ are $d$-separated given another variable $Z$ in a DAG representation then $X$ and $Y$ are conditionally independent given $Z$ in the corresponding distribution of the variables. However, \citet{spirtes-uai95,neal-jair00} show that without any specific assumption regarding the nature of dependence (i.e. linear, polynomial), the $d$-separation relations are not sufficient to entail all the corresponding conditional independence relations in a DCG. In a recent work, an alternative formulation called $\sigma$-separation is introduced which holds for a very general graphical setting \citep{forre-arxiv17}. 

Here, we consider a simplified version of the formal definition of $\sigma$-separation:

\begin{definition} [$\sigma$-separation] \citep{forre-arxiv17}

A walk $\langle v_0 . . . v_n \rangle$ in DCG $G = \langle \mathcal{V}, \mathcal{E} \rangle$ is $\sigma$-blocked by $C \subseteq V$ if:

\begin{enumerate}[parsep=0pt]
    \item its first node $v_0 \in C$ or its last node $v_n \in C$, or
    \item it contains a collider $v_k \notin AN_\mathcal{G}(C)$, or
    \item it contains a non-collider $v_k \in C$ that points to a node on the walk in another strongly connected component (i.e., $v_{k-1} \rightarrow v_k \rightarrow v_{k+1}$ with $v_{k+1} \notin SC_\mathcal{G}(v_k)$, $v_{k-1} \leftarrow v_k \leftarrow v_{k+1}$ with $v_{k-1} \notin SC_\mathcal{G}(v_k)$
    or $v_{k-1} \leftarrow v_k \rightarrow v_{k+1}$ with $v_{k-1} \notin SC_\mathcal{G}(v_k)$
    or $v_{k+1} \notin SC_\mathcal{G}(v_k)$).
\end{enumerate}

\noindent
If all paths in $\mathcal{G}$ between any node in set $A \subseteq \mathcal{V}$ and any node in set $B \subseteq \mathcal{V}$ are $\sigma$-blocked by a set $C \subseteq \mathcal{V}$, we say that $A$ is $\sigma$-separated from $B$ by $C$, and we write $A \overset{\sigma}{\underset{\mathcal{G}}{\indep}} B | C$.

\end{definition}


\subsection{$\sigma$-faithfulness}
$\sigma$-\textit{faithfulness} refers to the property which states that all statistical dependencies found in the distribution generated by a given causal structure model is entailed by the $\sigma$-separation relationships.

\begin{definition} [$\sigma$-faithfulness]
Given $\mathcal{X}_A$, $\mathcal{X}_B$, $\mathcal{X}_C$ as the distributions of variables $A$, $B$, $C$ respectively in solution $\mathcal{X}$ of a causal model $\mathcal{M}$, $\sigma$-\textit{faithfulness} states that if $\mathcal{X}_A$ and $\mathcal{X}_B$ are conditionally independent given $\mathcal{X}_C$, then $A$ and $B$ are $\sigma$-separated by $C$ in the corresponding possibly cyclic graphical model $\mathcal{G}$ of $\mathcal{M}$.
\end{definition}

\section{Relational Causal Models (RCMs)}


We adopt the definition of relational causal model used by previous work on relational causal discovery \citep{maier-uai13, lee-uai20}. We denote random variables and their realizations with uppercase and lowercase letters respectively, and bold to denote sets. We use a simplified Entity-Relationship model to describe relational data following previous work ~\citep{heckerman-isrl07}. A relational schema $\mathcal{S} = \langle \bm{\mathcal{E}}, \bm{\mathcal{R}}, \bm{\mathcal{A}}, card \rangle$ represents a relational domain where $\bm{\mathcal{E}}$, $\bm{\mathcal{R}}$ and $\bm{\mathcal{A}}$ refer to the set of entity, relationship and attribute classes respectively. It includes a cardinality function that constrains the number of times an entity instance can participate in a relationship. Figure \ref{fig:rcm} shows an example relational model that describes a simplified user-media engagement system. The model consists of three entity classes (User, Post, and Media), and two relationship classes (Reacts and Creates). Each entity class has a single attribute. The cardinality constraints are shown with crow’s feet notation— a user can react to multiple posts, multiple users can react to a post, and only a single media entity can create a post.

\sloppy A \textit{relational skeleton} $s$ is 
an instantiation of a relational schema $\mathcal{S}$, represented by an undirected graph of entities and relationships. Figure \ref{sfig:skel} shows an example skeleton of the relational model from Figure \ref{fig:rcm}. It shows that Alice and Bob both react to post P1. Alice also reacts to post P2. P1 and P2 both are created by media M1. There could be infinitely many possible skeletons for a given RCM. We  denote the set of all skeletons for schema $\mathcal{S}$ as $\sum_\mathcal{S}$. 

Given a relational schema, we can specify relational paths, which intuitively correspond to ways of traversing the schema. For the schema shown in Figure \ref{fig:rcm}, possible paths include [User, Reacts, Post] (the posts a user reacts to), as well as [User, Reacts, Post, Reacts, User] (other users who react to the same post). \textit{Relational variables} consist of a relational path and an attribute. For example, the relational variable [User, Reacts, Post].Engagement corresponds to the overall engagement of the post that a user reacts to. The first item (i.e. $User$) in the relational path corresponds to the \textit{perspective} of the relational variable. A terminal set, $P|_{i_k}$ is the terminal item on the relational path $P = [I_j, . . . , I_k]$ consisting of instances of class $I_k \in \bm{\mathcal{E}} \cup \bm{\mathcal{R}}$.

A relational causal model $\mathcal{M} = \langle \mathcal{S},\mathcal{D} \rangle$, is a collection of relational dependencies defined over schema $\mathcal{S}$. \textit{Relational dependencies} consist of two relational variables, cause and effect. As an example, consider the following relational dependency [Post, Reacts, User].Sentiment $\rightarrow$ [Post].Engagement which states that the engagement of a post is affected by the actions of users who react to that post. In Figure \ref{fig:rcm}, the arrows represent relational dependencies. Note that, all causal dependencies are defined with respect to a specific perspective. A relational model $\mathcal{M} = (\mathcal{S}, \mathcal{D})$ is said to be cyclic if the set of relational dependencies $\mathcal{D}$ constructs one or more directed cycles of arbitrary length. There is a direct feedback loop in the relational model of Figure \ref{fig:rcm} making it a cyclic relational causal model.

\subsection{Ground Graph and $\sigma$-Abstract Ground Graph}
\label{sec:gg_agg}

\sloppy
A realization of a relational model $\mathcal{M}$ with a relational skeleton is referred to as the \textit{ground graph} $GG_\mathcal{M}$. It is a directed graph consisting attributes of entities in the skeleton as nodes and relational dependencies among them as edges. A single relational model is actually a template for a set of possible ground graphs based on the given schema. A ground graph has the same semantic as a graphical model. Given a relational model $\mathcal{M}$ and a relational skeleton $s$, we can construct a ground graph \ggms{} by applying the relational dependencies as specified in the model to the specific instances of the relational skeleton. Figure \ref{sfig:gg} shows the ground graph for the relational model from Figure \ref{fig:rcm}. The relational dependencies present in the given RCM may temp one to conclude a conditional independence statement: \textit{[User].\mstance $\,\indep\,$ [Media].Preference | [Post].Engagement}. However, when the model is unrolled in a ground graph we see the corresponding statement is not true (i.e. \textit{[Bob].\mstance $\,\nindep$ [M1].Preference | [P1].Engagement}) since there is an alternative path through \textit{[Alice].\mstance} and \textit{[P2].Engagement} which is activated when conditioned on \textit{[P1].Engagement}. It shows why generalization over all possible ground graphs is hard.

A $\sigma$-\textit{abstract ground graph} ($\sigma$-AGG) is an abstract representation that solves the problem of generalization by capturing the consistent dependencies in all possible ground graphs and representing them as a directed graph. $\sigma$-AGGs are defined for a specific perspective and \textit{hop threshold}, $h$. Hop threshold refers to the maximum length of the relational paths allowed in a specific $\sigma$-AGG. There are two types of nodes in $\sigma$-AGG, relational variables, and intersection variables. Intersection variables are constructed from pairs of relational variables with non-empty intersections~\citep{maier-arxiv13}. For example, [User, Reacts, Post] refers to the set of posts a user reacts to whereas [User, Reacts, Post, Reacts, User, Reacts, Post] refers to the set of other posts reacted to by other users who also reacted to the same post as the given user. These two sets of posts can overlap which is reflected by the corresponding intersection variable. Edges between a pair of nodes of $\sigma$-AGG exist if the instantiations of those constituting relational variables contain a dependent pair in all ground graphs. Figure \ref{sfig:agg} presents the $\sigma$-AGG from the perspective of $User$ and with $h = 6$ corresponding to the model from Figure \ref{fig:rcm}. The $\sigma$-AGG shows that the sentiment of a user is no longer independent of media preference given just engagements of the corresponding posts the user reacts to. We also need to condition on the sentiment of other users who reacted to the same post.

\subsection{Relational $\sigma$-separation}
Conditional independence facts are only useful when they hold across all ground graphs that are consistent with the model. \citet{maier-arxiv13} show that relational $d$-separation is sufficient to achieve that for acyclic models. However, such abstraction is not possible for cyclic models since the correctness of $d$-separation is not guaranteed for cyclic graphical models for the general form of dependency~\citep{spirtes-uai95, neal-jair00}. In recent work, \citep{ahsan-clear22} introduced relational $\sigma$-separation criteria specifically for cyclic relational models:

\begin{definition} [Relational $\sigma$-separation]
Let $\bm{X}$, $\bm{Y}$, and $\bm{Z}$ be three distinct sets of relational variables with the same perspective $B \in \bm{\mathcal{E}} \cup \bm{\mathcal{R}}$ defined over relational schema $\mathcal{S}$. Then, for relational model structure $\mathcal{M}$, $\bm{X}$ and $\bm{Y}$ are $\sigma$-separated by $\bm{Z}$ if and only if, for all skeletons $s \in \sum_\mathcal{S}$, $\bm{X}|_b$ and $\bm{Y}|_b$ are $\sigma$-separated by $\bm{Z}|_b$ in ground graph $GG_{\mathcal{M}_s}$ for all instances $b \in s(B)$ where $s(B)$ refers to the instances of $B$ in skeleton $s$.
\end{definition}

The definition directly follows from the definition of relational $d$-separation. If there exists even one skeleton and faithful distribution represented by the relational model for which $\bm{X} \not\!\perp\!\!\!\perp \bm{Y} | \bm{Z}$, then $\bm{X}|_b$ and $\bm{Y}|_b$ are not $\sigma$-separated by $\bm{Z}|_b$ for $b \in s(B)$.

\section{Relational Causal Discovery (RCD)}

The RCD algorithm developed by \citet{maier-uai13} is the first sound and complete algorithm that can discover the abstract ground graph of a relational causal model (RCM) under the assumption of $d$-faithfulness, sufficiency, acyclicity, and a maximum hop threshold $h$. It is designed based on the PC algorithm with some additional steps introduced specifically to handle relational aspects of the representation. Similar to the PC algorithm, the steps of RCD are divided into two phases: 1) skeleton detection and 2) edge orientation. The first phase is identical to the PC algorithm. The second phase is also inspired by PC and uses all four orientation rules from the PC algorithm. An additional important step that RCD performs is the propagation of edge orientation which refers to the idea of orienting all edges associated with a certain relational dependency in all possible AGGs. This helps ensure completeness as well as reduces redundant iteration of the algorithm. Moreover, RCD introduces Relational Bivariate Orientation (RBO)- an orientation rule specifically designed for relational models. It applies to unshielded triples with end nodes having the same attribute type and entity type. Unlike \textit{collider detection} in PC, RBO can orient an unshielded triple to both a collider and a fork structure which yields significantly more orientations than using only the four PC orientation rules. \citet{maier-uai13} provides theoretical guarantees for soundness and completeness of RCD.

\subsection{Demonstration on Real Data}

\begin{figure}[!ht]
    \centering

    \includegraphics[width=.40\textwidth]{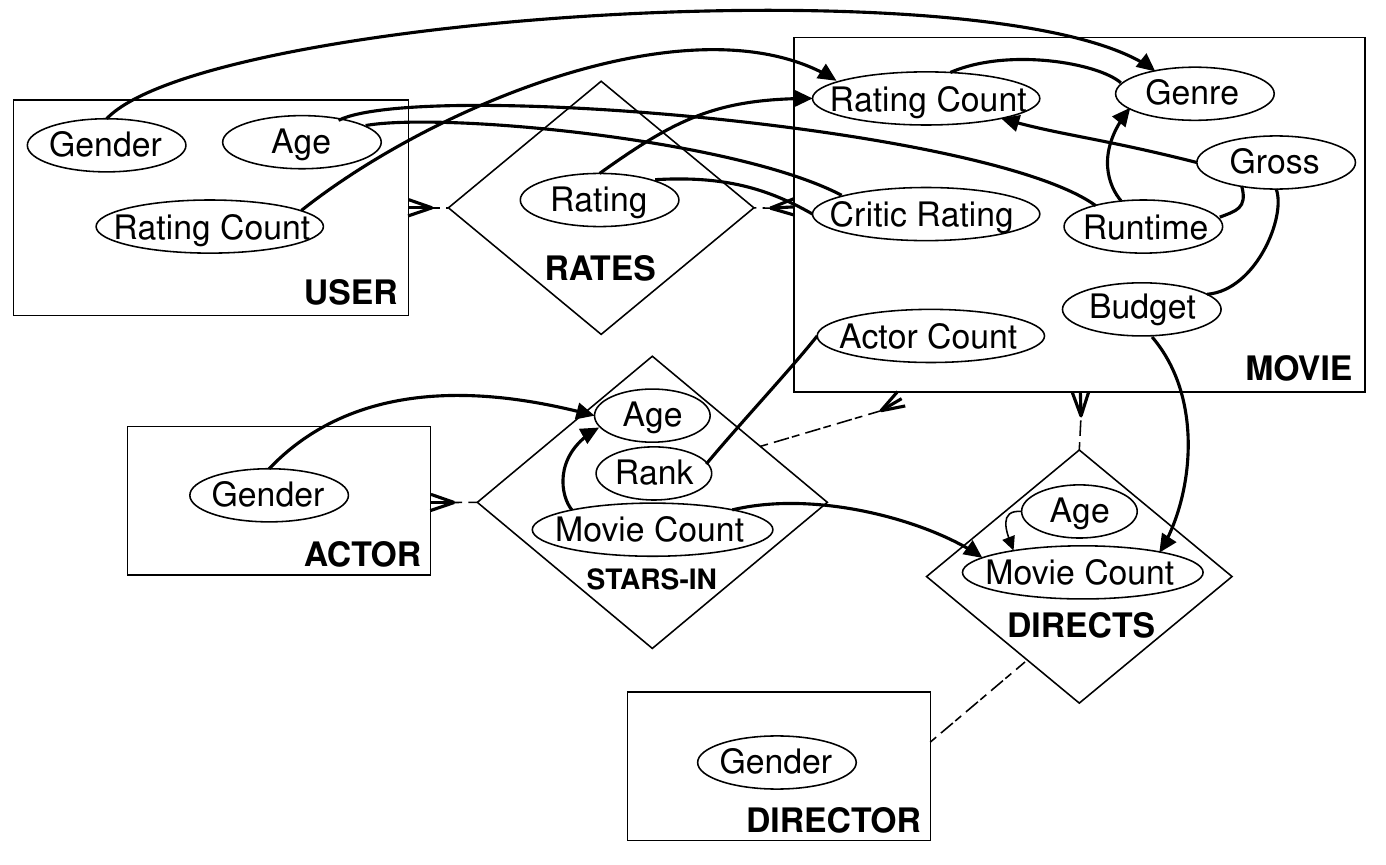}
    
    \caption{
        RCD-learned model of MovieLens+~\cite{maier-uai13}.
     }
    \label{fig:demo_rcd}
\end{figure}

\citet{maier-uai13} applied RCD to the MovieLens+ database, a combination of the UMN MovieLens database (www.grouplens.org); box office, director, and actor information collected from IMDb (www.imdb.com); and average critic ratings from Rotten Tomatoes (www.rottentomatoes.com). They used a sample of 75,000 ratings. For testing conditional independence they used the significance of coefficients in linear regression and considered the average aggregation function for relational variables. They ran RCD with a hop threshold of 4, maximum depth of 3, and an effect size threshold of 0.01. The RCD-generated output is given in Figure \ref{fig:demo_rcd}.

\section{Proofs}

\begin{theorem}
Considering Assumption \ref{asm:sfaith}, \ref{asm:rel_acy}, \ref{asm:sagg} and causal sufficiency holds, RCD is 

\begin{enumerate}[(i)]
    \item sound: for all \sagg{}s $G$, \mr{G} contains $G$;
    \item arrowhead complete: for all \sagg{}s $G$: if $i \notin AN_{\tilde{G}}(j)$ for any DCG $\tilde{G}$ that is $\sigma$-Markov equivalent to $G$, then there is an arrowhead $j\; \stararrow \;i$ in \mr{G} 
    \item tail complete: for all \sagg{}s $G$, if $i \in AN_{\tilde{G}}(j)$ 
    in any DCG $\tilde{G}$ that is $\sigma$-Markov equivalent to $G$, 
    then there is a tail $i \rightarrow j$ in \mr{G};
    \item Markov complete: for all \sagg{}s $G_1$ and $G_2$, $G_1$ is $\sigma$-Markov equivalent to $G_2$ iff $\mr{G_1} = \mr{G_2}$
\end{enumerate}
in the $\sigma$-separation setting given sufficient hop threshold.
\end{theorem}

\begin{proof}
    To prove soundness, let $G$ be a \sagg{} and $\mathcal{P} = \mr{G}$. The acyclic soundness of RCD means that for all AGGs $G^\prime$, $\mr{G^\prime}$ contains $G^\prime$. Hence, by Definition \ref{dfn:rel_acy} and Assumption \ref{asm:rel_acy}, $\mathcal{P}$ contains $G^\prime$ for all acyclifications $G^\prime$. But then $\mathcal{P}$ must contain $G$:

    \begin{itemize}
        \item if two vertices $i$, $j$ are adjacent in $\mathcal{P}$ then there is an inducing path between $i$, $j$ in any acyclification of $G$, which holds if and only if there is an inducing path between $i$, $j$ in $G$ (Proposition \ref{prop:acy}(iii));
        
        \item if $i\; \stararrow j$ in $\mathcal{P}$, then $j \notin AN_{G^\prime} (i)$ for any acyclification $G^\prime$ of $G$, and hence $j \notin AN_G(i)$ (Proposition  \ref{prop:acy}(i));
        
        \item if $i \rightarrow j$ in $\mathcal{P}$, then $i \in AN_{G^\prime}(j)$ for all acyclifications $G^\prime$ of $G$, and hence $i \in AN_G(j)$ (Proposition  \ref{prop:acy}(ii)).
    \end{itemize}

    To prove arrowhead completeness, let $G$ be a \sagg{} and suppose that $i \notin AN_{\tilde{G}}(j)$ in any DCG $\tilde{G}$ that is $\sigma$-Markov equivalent to $G$. Since $G^\prime$ is $\sigma$-Markov equivalent to $G$, this implies in particular that for all AGGs $\tilde{G}$ that are $d$-Markov equivalent to $G^\prime$, $i \notin AN_{\tilde{G}}(j)$. Because of the acyclic arrowhead completeness of RCD, there must be an arrowhead $j\; \stararrow \;i$ in $ \mr{G^\prime} = \mr{G}$. 
    
    To prove tail completeness, let $G$ be a \sagg{} and suppose that $i \in AN_{\tilde{G}}(j)$ in any DCG $\tilde{G}$ that is $\sigma$-Markov equivalent to $G$. Since $G^\prime$ is $\sigma$-Markov equivalent to $G$, this implies in particular that for all AGGs $\tilde{G}$ that are $d$-Markov equivalent to $G^\prime$, $i \in AN_{\tilde{G}}(j)$. Because of the acyclic tail completeness of RCD, there must be a tail $i \rightarrow j$ in $ \mr{G^\prime} = \mr{G}$.
    
    To prove Markov completeness: Definition \ref{dfn:rel_acy} and Proposition \ref{prop:im} imply both $\im{G_1} = \im[d]{G_1^\prime}$ and $\im{G_2} = \im[d]{G_2^\prime}$. From the acyclic Markov completeness of RCD\footnote{Since relational d-separation is equivalent to the Markov condition and it is sound and complete on abstract ground graph~\cite{maier-arxiv13}}, it then follows that $G_1^\prime$ must be $d$-Markov equivalent to $G_2^\prime$, and hence $G_1$ must be $\sigma$-Markov equivalent to $G_2$.

\end{proof}

\end{document}